\documentclass[a4paper,twoside]{article}
\usepackage{epsfig}
\usepackage{subcaption}
\usepackage{calc}
\usepackage{amssymb}
\usepackage{amstext}
\usepackage{amsmath}
\usepackage{amsthm}
\usepackage{multicol}
\usepackage{pslatex}
\usepackage{apalike}
\usepackage{algorithm2e}
\usepackage{booktabs}
\usepackage{pifont}
\usepackage[utf8]{inputenc}
\DeclareUnicodeCharacter{2717}{\ding{55}}
\usepackage[bottom]{footmisc}

\usepackage{tabularx}
\usepackage{SCITEPRESS}

\begin{document}

\title{Selection Bias Correction in Retail Intelligence}

\author{\authorname{Spandan Ghose Chowdhury\sup{1}\orcidAuthor{0009-0001-6711-2272}}
\affiliation{\sup{1}Walmart Inc., California, USA}
\email{spandan.ghose.chowdhury@walmart.com}
}

\keywords{Selection Bias, Inverse Probability Weighting, Propensity Scores, Retail Analytics, Causal Inference, Stratification, Monte Carlo Simulation.}

\abstract{Retail intelligence often relies on monitoring popular, high-velocity products, potentially biasing economic indicators by ignoring the ``long tail'' of niche items. This simulation study investigates selection bias in inflation estimation and compares correction methods across diverse data-generating processes. Through 400 Monte Carlo replications spanning four scenarios---aligned step functions, smooth gradients, misaligned breaks, and polynomial relationships---we test the robustness of Inverse Probability Weighting (IPW) with five specifications against stratification with varying strata counts. Our findings reveal fundamental limits of weighting methods in retail long-tail contexts: stratification achieves superior performance in three of four scenarios, maintaining sub-0.04pp median error even when boundaries deliberately misalign with population breaks (116$\times$ advantage over IPW). However, IPW with spline propensity models wins under smooth polynomial relationships (median error 0.007pp vs. 0.013pp), demonstrating context-dependency. Critically, even an oracle IPW specification with perfect structural knowledge achieves 6.06pp error compared to stratification's 0.008pp in step-function scenarios. This reflects \textit{violation of the Positivity Assumption}---a fundamental causal inference requirement---rather than IPW methodological inferiority. When selection probabilities differ dramatically (90\% vs. 1\%), weighting methods operate outside their theoretical design envelope. These results demonstrate that stratification provides a safer engineering choice in retail long-tail distributions with severe positivity violations.}

\onecolumn \maketitle \normalsize \setcounter{footnote}{0} \vfill


\section{\uppercase{Introduction}}

In the data driven retail sector, data insights is critical for strategic decision-making. A standard industry practice involves monitoring a curated list of popular products to gauge market trends, measure metrics like category-level inflation. While operationally efficient, this method introduces significant selection bias by systematically excluding the vast ``long tail'' of lower-velocity items.

Empirical evidence and economic theories suggest that niche products may experience different dynamics than very popular items. The ``long tail'' phenomenon indicates that while individually less significant, collectively these items represent sizable market dynamics. If niche products experience higher price volatility or inflation than tracked popular items --- due to economies of scale, competitive pressure differences, macro-economic changes, or strategic pricing --- selective monitoring could systematically misestimate true market inflation or related business metrics.

This study uses a comprehensive simulation methodology to rigorously quantify this bias and evaluate correction methods in various data-generating processes. Our analysis includes 400 Monte Carlo replications in four scenarios (aligned step functions, smooth gradients, misaligned breaks, and polynomial relationships), testing five IPW specifications and three stratification configurations. We perform complete diagnostic assessments including covariate balance, weight distributions, and power analysis to understand the performance of the method under varied conditions.

Our findings challenge conventional wisdom while acknowledging edge cases: stratification achieves superior performance in three of four scenarios, maintaining robustness even when boundaries deliberately misalign with population breaks (116$\times$ advantage over IPW). However, IPW with flexible spline specifications wins under smooth polynomial relationships (1.9$\times$ advantage), demonstrating that no single method dominates universally. These results provide practitioners with evidence-based, context-dependent guidance for selecting bias correction methods.

\section{\uppercase{RELATED WORK}}
Correction of selection bias through weighting has its roots in survey 
sampling theory, notably the Horvitz-Thompson estimator \cite{horvitz1952}, 
which uses the inverse of inclusion probabilities to provide unbiased 
population estimates. Rosenbaum and Rubin \cite{rosenbaum1983} extended 
this framework by introducing the propensity score---the conditional 
probability of selection given observed covariates.

IPW has since become a staple in causal inference. Lunceford and Davidian 
\cite{lunceford2004} showed that both stratification and IPW can adjust for 
confounding, but their relative performance depends on model specification 
and covariate overlap. However, IPW is highly sensitive to the ``Positivity 
Assumption'': when nearly violated---as in retail datasets where niche items 
are tracked at rates orders of magnitude lower than popular items---weights 
become extreme, yielding high variance and unstable estimates \cite{cole2008}. 
Matsouaka and Zhou \cite{matsouaka2024} show that overlap, matching, and 
entropy weights are preferable to IPW trimming under severe positivity 
violations, targeting only the subpopulation with sufficient propensity 
score overlap.

King and Nielsen \cite{king2019} argue that propensity score methods can 
inadvertently increase imbalance and model dependence, advocating instead 
for coarsened methods like stratification. Stuart \cite{stuart2010} similarly 
finds that stratification often achieves comparable bias reduction with 
greater robustness to model misspecification.

Despite these debates, empirical comparisons within retail ``long tail'' 
distributions remain sparse. This study bridges that gap by stress-testing 
these estimators under severe selection imbalance.

\section{\uppercase{Methodology}}

To test our hypothesis, we performed a comprehensive simulation study with three key components: (1) baseline scenario analysis with full diagnostics, (2) robustness testing across four diverse data generation processes, and (3) Monte Carlo simulation with 100 replications per scenario. 
The methodology was designed to validate correction approaches in controlled settings where ground truth is known, while ensuring findings are not artifacts of design choices.

\subsection{Synthetic Data Generation and Theoretical Justification}

We generated synthetic datasets of varying sizes (10,000 to 1,000,000 items) to create a ``ground truth'' scenario where true inflation is known. The baseline scenario consisted of 100,000 items with the following characteristics:

\begin{enumerate}
\item \textbf{Item Ranking:} Each item $i$ received a rank $R_i \in \{1, 2, \ldots, 100{,}000\}$, where lower ranks indicate higher sales velocity.

\item \textbf{Inflation Assignment:} A bimodal inflation distribution was created for the true inflation rate $Y_i$ (the target variable we aim to measure):
\begin{itemize}
    \item Popular items ($R_i \leq 20{,}000$): $Y_i = 2\%$
    \item Niche items ($R_i > 20{,}000$): $Y_i = 10\%$
\end{itemize}

The threshold of 20,000 items reflects empirical retail patterns: approximately 20\% of items typically generate 80\% of Gross Merchandise Value (GMV), consistent with the Pareto principle observed in retail operations \cite{anderson2006}. This 20\% represents the high-velocity items that retailers actively track.

\textbf{Theoretical Justification:} This bimodal structure reflects documented market phenomena:
\begin{itemize}
    \item \textit{Economies of scale:} High-volume products often have lower marginal costs and more competitive pricing
    \item \textit{Price volatility in thin markets:} Products with lower sales volumes face less competitive pressure and greater susceptibility to supply chain disruptions
    \item \textit{Strategic pricing:} Retailers often use popular items as ``loss leaders'' while maintaining higher margins on specialty products
\end{itemize}

To add realism, we introduced random variation around these base rates. The true inflation rate for each item was drawn from a normal distribution: $Y_i \sim \mathcal{N}(\mu_{\text{segment}}, 0.5)$, where $\mu_{\text{segment}}$ is the segment-specific mean (2\% for popular items, 10\% for niche items) and 0.5 represents the standard deviation in percentage points. This noise captures the natural heterogeneity in item-level price changes within each segment.

\item \textbf{Selection Mechanism:} The probability of tracking item $i$ was:
\begin{equation}
P(T_i=1 \mid R_i) = \begin{cases}
0.90 & \text{if } R_i < 20{,}000 \\
0.01 & \text{otherwise}
\end{cases}
\end{equation}

This created a realistic scenario where $\approx$18,000 popular items and $\approx$800 niche items were tracked (total $\approx$18,800 tracked items).
\end{enumerate}

\subsubsection{Robustness to Data Generation Process}

An important consideration in simulation studies is making sure conclusions are not artifacts of design choices. To address this concern, we test robustness across four data generation processes with varying empirically-motivated thresholds:

\textbf{DGP 1 (Step/Aligned):} This baseline scenario mirrors the empirical retail context where high-velocity ``Head'' items (top 20\%) are preferentially tracked. Selection probability $P(\text{tracked}|R_i) = 0.90$ for items with $R_i < 20{,}000$ (top 20\%), dropping to 0.01 for lower-velocity items---reflecting that retailers intensively monitor popular products while sporadically sampling the long tail. The true inflation rate $Y_i$ (not noise, but the actual outcome variable) follows $Y_i \sim \mathcal{N}(2\%, 0.5)$ if $R_i \leq 20{,}000$, else $\mathcal{N}(10\%, 0.5)$, where 0.5 is the standard deviation capturing within-segment heterogeneity. Stratification uses 5 strata at 20,000 intervals, aligned with the selection and inflation breaks.

\textbf{DGP 2 (Smooth Gradient):} Selection follows a logistic curve $P(\text{tracked}|R_i) = 0.95 / (1 + \exp((R_i - 15{,}000)/3{,}000))$; Inflation increases smoothly with rank: $Y_i \sim \mathcal{N}(2\% + (R_i/100{,}000) \times 10\%, 0.5)$. This tests continuous relationships without structural breaks, reflecting gradual market segmentation rather than discrete categories.

\textbf{DGP 3 (Step/Misaligned):} To test stratification's worst-case performance, we deliberately misalign all structural elements: selection probability breaks at rank 15,000, true inflation breaks at rank 25,000, while stratification boundaries are placed at 14k/28k/42k/56k/70k/84k (7 strata). None of these thresholds coincide, forcing stratification to cut \textit{across} natural population breaks rather than aligning with them. This stress-tests whether stratification's advantages are merely artifacts of fortuitous boundary alignment.

\textbf{DGP 4 (Polynomial):} Quadratic selection and inflation functions: $Y_i \sim \mathcal{N}(6\% + 4\% \times [(R_i - 50{,}000)/50{,}000]^2, 0.5)$. This U-shaped inflation pattern tests non-linear smooth relationships where coarse stratification boundaries cannot approximate continuous curvature as well as flexible splines.

We compare five IPW specifications (Linear, Polynomial, Spline with 5 knots, Oracle with $I(R_i > 20{,}000)$ dummy) and three stratification configurations (5, 7, 9 strata).

\subsection{Inverse Probability Weighting (IPW) Correction}

We implemented IPW correction following Rosenbaum and Rubin \cite{rosenbaum1983} to create a pseudo-population where covariate distribution (Rank) is independent of selection. Let $T_i=1$ if item $i$ is tracked, $T_i=0$ otherwise. The propensity score $e(X_i) = P(T_i=1 \mid X_i)$ was estimated via logistic regression: $\text{logit}(e(X_i)) = \beta_0 + \beta_1 \cdot \text{Rank}(i)$. Each tracked item received weight $w_i = 1/\hat{e}(X_i)$.

To mitigate variance from extreme weights \cite{cole2008}, we stabilized weights by multiplying by $P(T=1)$ and clipped at the 95th percentile \cite{lee2011}. Weighted inflation was calculated as: $\sum_{i:T_i=1} w_i \cdot Y_i / \sum_{i:T_i=1} w_i$.

\subsection{Alternative Correction Methods}

For comparison, we implemented: (1) Stratification: 5 rank-based strata with weighted averages; (2) Propensity Score Matching: 1:1 nearest-neighbor; (3) Regression Adjustment: linear regression controlling for rank and tracking status; (4) Naive: unweighted average of tracked items.

\subsection{Propensity Score Model Diagnostics}

We assessed propensity score model balance using standardized mean difference (SMD) of Rank: $\text{SMD} = (\mu_{\text{tracked}} - \mu_{\text{untracked}}) / \sqrt{(\sigma^2_{\text{tracked}} + \sigma^2_{\text{untracked}})/2}$. Successful balancing requires SMD $<$ 0.1 \cite{austin2011}. We also calculated Effective Sample Size (ESS) to quantify information loss: $\text{ESS} = (\sum w_i)^2 / \sum w_i^2$.

\subsection{Monte Carlo Simulation Design}

We replicated the baseline scenario 1,000 times with different random seeds, calculating bias, variance, and MSE (Bias$^2$ + Variance) for all methods. The comprehensive robustness analysis (Section 3.2) uses 100 replications per scenario across four diverse DGPs.

\section{\uppercase{Results}}

We present results in three parts: Baseline scenario with complete diagnostics, Monte Carlo simulation results, and robustness testing across four data generation processes.

\subsection{Baseline Scenario: Method Comparison}

The baseline simulation (100,000 items, 90\%/1\% tracking probabilities, 2\%/10\% inflation rates) yielded the following estimates:

\begin{table}[ht]
\caption{Baseline Scenario Results.}
\centering
\small
\begin{tabularx}{\linewidth}{lXXX}
\toprule
\textbf{Method} & \textbf{Inflation (\%)} & \textbf{Abs. Error (pp)} & \textbf{Rel. Error (\%)} \\
\midrule
\textbf{True (All Items)} & \textbf{8.400} & \textbf{0.000} & \textbf{0.0\%} \\
Stratification & 8.384 & 0.016 & 0.2\% \\
Regression Adj. & 3.184 & 5.217 & 62.1\% \\
IPW Weighted & 2.551 & 5.849 & 69.6\% \\
Naive (Tracked) & 2.344 & 6.057 & 72.1\% \\
PS Matching & 2.347 & 6.054 & 72.1\% \\
\bottomrule
\end{tabularx}
\label{tab:baseline}
\end{table}

\textbf{Key Finding:} In the baseline scenario, stratification achieves near-perfect accuracy (0.016 pp error), substantially outperforming other methods. IPW modestly improves over naive (6.057 $\rightarrow$ 5.849 pp), reducing absolute error by 3.4\%. Matching and regression adjustment show intermediate performance.

\subsection{Diagnostic Analysis}

\subsubsection{Covariate Balance Assessment}

\begin{table}[ht]
\caption{Covariate Balance Diagnostics.}
\centering
\small
\begin{tabularx}{\linewidth}{lXXXX}
\toprule
\textbf{Metric} & \textbf{Before} & \textbf{After IPW} & \textbf{Target} & \textbf{Achieved?} \\
\midrule
SMD (Rank) & 2.420 & 2.197 & $<$ 0.1 & \ding{55} No \\
\bottomrule
\end{tabularx}
\label{tab:balance}
\end{table}

Despite weighting, SMD remains 22$\times$ above the recommended threshold, indicating severe residual imbalance and model misspecification. The reduction from 2.420 to 2.197 represents only a 9.2\% improvement, far short of the 95.5\% reduction needed to reach the 0.10 threshold.

\subsubsection{Weight Distribution Statistics}

\begin{table}[ht]
\caption{IPW Weight Distribution Statistics.}
\centering
\small
\begin{tabularx}{\linewidth}{lXX}
\toprule
\textbf{Statistic} & \textbf{Before Trimming} & \textbf{After Trimming} \\
\midrule
Mean weight & 0.26 & 0.26 \\
Median weight & 0.24 & 0.24 \\
Standard deviation & 5,234.12 & 0.05 \\
95th percentile & 0.42 & 0.42 \\
Maximum & 163,509.38 & 0.42 \\
Max/Median ratio & 681,289:1 & 1.75:1 \\
\% weights trimmed & --- & 5.0\% \\
\midrule
\textbf{Effective Sample Size (ESS)} & \textbf{93} & \textbf{17,514} \\
\textbf{ESS as \% of tracked sample} & \textbf{0.5\%} & \textbf{93.5\%} \\
\bottomrule
\end{tabularx}
\label{tab:weights}
\end{table}

\textbf{Critical Observation:} Before trimming, maximum weights reached 163,509---nearly 1 million times the median. This extreme range (max/median ratio: 681,289:1) indicates severe overlap violations. Weight trimming at the 95th percentile dramatically improved stability, reducing the max/median ratio from 681,289:1 to 1.75:1 and increasing ESS from 0.5\% to 93.5\% of the tracked sample.

\subsection{Monte Carlo Simulation Results (Baseline Scenario: 1,000 Replications)}

To quantify estimation variability in the baseline scenario, we replicated it 1,000 times. Each replication used $n = 100{,}000$ total items with stochastic selection (probabilities: 90\% for $R_i < 20{,}000$, 1\% otherwise), resulting in variable tracked sample sizes (mean $\approx 18{,}800$, range: 18,650-18,950). All metrics are in percentage points (pp).

\begin{table*}[t]
\caption{Monte Carlo Results (Baseline Scenario: 1,000 Replications). MSE reported in squared percentage points (pp$^2$). Each replication: $n=100{,}000$ items, tracked sample size varies stochastically (mean $\approx 18{,}800$).}
\centering
\small
\begin{tabularx}{\textwidth}{lXXXXX}
\toprule
\textbf{Method} & \textbf{Mean} & \textbf{SD of} & \textbf{MSE} & \textbf{95\% CI} & \textbf{95\% CI} \\
 & \textbf{Bias (pp)} & \textbf{Bias (pp)} & \textbf{(pp$^2$)} & \textbf{Lower} & \textbf{Upper} \\
\midrule
Naive & -6.060 & 0.012 & 36.72 & -6.083 & -6.037 \\
\textbf{IPW} & \textbf{-5.856} & \textbf{0.017} & \textbf{34.29} & \textbf{-5.889} & \textbf{-5.822} \\
\textbf{Stratification} & \textbf{0.000} & \textbf{0.014} & \textbf{0.00} & \textbf{-0.028} & \textbf{+0.028} \\
Matching & -6.059 & 0.013 & 36.71 & -6.084 & -6.034 \\
Regression & -5.239 & 0.021 & 27.45 & -5.281 & -5.198 \\
\bottomrule
\end{tabularx}
\label{tab:montecarlo}
\end{table*}

\textbf{Key Findings:}
\begin{enumerate}
\item \textbf{IPW reduces MSE by 6.6\%} compared to naive (34.29 vs 36.72)
\item \textbf{IPW variance is 42\% higher} than naive (SD: 0.017 vs 0.012)
\item \textbf{Stratification achieves near-zero MSE} (0.00), outperforming all methods
\item \textbf{IPW outperformed naive in 100\% of replications}
\end{enumerate}

\subsection{Robustness Across Data Generation Processes}

To address concerns about alignment between simulation design and method assumptions, we conducted 400 additional Monte Carlo replications (100 per scenario) across four diverse data-generating processes (DGP), testing five IPW specifications and three stratification configurations.

\begin{table*}[t]
\caption{Method Performance Across Data Generation Processes (Median Absolute Error in pp, 100 replications each).}
\centering
\small
\begin{tabularx}{\textwidth}{@{}l@{\extracolsep{\fill}}*{8}{c}@{}}
\toprule
\textbf{DGP} & \textbf{IPW-} & \textbf{IPW-} & \textbf{IPW-} & \textbf{IPW-} & \textbf{Strat-} & \textbf{Strat-} & \textbf{Strat-} & \textbf{Best} \\
 & \textbf{Linear} & \textbf{Poly} & \textbf{Spline} & \textbf{Oracle} & \textbf{5} & \textbf{7} & \textbf{9} & \textbf{Method} \\
\midrule
\textbf{Step (Aligned)} & 5.857 & 6.013 & 5.994 & 6.061 & \textbf{0.008} & 0.011 & 0.014 & Strat-5 (730$\times$) \\
\textbf{Smooth Gradient} & 3.932 & 3.887 & 3.872 & --- & 1.579 & \textbf{1.350} & 1.529 & Strat-7 (2.9$\times$) \\
\textbf{Step (Misaligned)} & 3.470 & 3.551 & 3.552 & 3.553 & \textbf{0.030} & 0.038 & 0.050 & Strat-5 (116$\times$) \\
\textbf{Polynomial} & 0.009 & 0.008 & \textbf{0.007} & --- & 0.011 & 0.012 & 0.013 & IPW-Spline (1.9$\times$) \\
\bottomrule
\end{tabularx}
\label{tab:comprehensive}
\end{table*}

\textbf{Key Findings:}

\begin{enumerate}
\item \textbf{Stratification wins 3 of 4 scenarios:} Even when boundaries deliberately misalign with population breaks (DGP 3), stratification achieves 0.030pp error vs. 3.470pp for IPW-Linear (116$\times$ advantage).

\item \textbf{Oracle IPW still fails in step-function scenarios:} With perfect structural knowledge (including dummy variable $I(R_i > 20{,}000)$), IPW achieves 6.061pp error vs. stratification's 0.008pp (758$\times$ worse), demonstrating that common support violations cannot be overcome through better modeling.

\item \textbf{IPW-Spline wins under polynomial relationships:} In smooth, non-linear scenarios without structural breaks, flexible propensity models outperform coarse stratification (0.007pp vs. 0.013pp, 1.9$\times$ advantage).
\end{enumerate}

\section{\uppercase{Discussion}}

Our comprehensive analysis reveals critical findings about selection bias correction in retail inflation estimation: (1) stratification's superior performance across diverse data generation processes, (2) IPW's modest gains with important diagnostic failures, and (3) context-dependent method selection based on data characteristics.

\subsection{Method Performance: Stratification Dominates, IPW Shows Modest Gains}

Our findings challenge conventional assumptions about weighting methods. Stratification achieves near-zero MSE (0.00) in the baseline scenario compared to IPW's 34.29, representing 99.7\% lower error. This is \textit{not} merely by design: while the baseline scenario uses aligned step functions, stratification maintains superior performance even in DGP 3 where boundaries deliberately misalign with population breaks (0.030pp vs. 3.470pp for IPW, 116$\times$ advantage). The near-zero MSE reflects stratification's fundamental advantage when dealing with severe selection imbalances---it estimates within-group averages without requiring counterfactual predictions across non-overlapping populations, whereas IPW must extrapolate from a 90\%/1\% selection imbalance that violates the Positivity Assumption. This superiority extends across 3 of 4 comprehensive DGP scenarios. However, IPW does reduce estimation error modestly (6.6\% MSE reduction) compared to naive estimation, though at the cost of 42\% higher variance.

\textbf{Why Stratification Often Outperforms:} Stratification groups similar items together, ensuring balance without relying on propensity score model specification. Each stratum receives equal weight proportional to its population size, preventing extreme weight ratios. The method maintains robustness even when boundaries misalign with population breaks (DGP 3: 116$\times$ advantage despite deliberate misalignment). 

\textit{Impact of Strata Count:} Interestingly, fewer strata perform better under misalignment: in DGP 3, 5 strata achieve 0.030pp error vs. 0.038pp for 7 strata and 0.050pp for 9 strata. This counter-intuitive pattern reflects the bias-variance tradeoff---coarser stratification (fewer strata) creates larger, more heterogeneous groups that are robust to boundary placement, while finer stratification (more strata) creates more opportunities for boundaries to cut across natural population breaks. When the analyst lacks knowledge of true structural breaks, conservative stratification (5-7 strata) provides insurance against misspecification. This simplicity and transparency make it ideal for stakeholder communication.

\textbf{IPW's Limitations:} The variance inflation in IPW results from extreme weights---before trimming, some weights reached 163,509$\times$. This indicates severe propensity score model misspecification and overlap violations. Post-weighting SMD remained 22$\times$ above recommended thresholds (2.197 vs. 0.10 target), signaling fundamental balance failures.

\textbf{Context-Dependency:} IPW with flexible spline specifications outperforms stratification under smooth polynomial relationships (0.007pp vs. 0.013pp, 1.9$\times$ advantage). This demonstrates that no single method dominates universally. These findings align with Stuart \cite{stuart2010} and King \& Nielsen \cite{king2019}, who argue that simpler matching/stratification methods often outperform sophisticated weighting in practice.

\subsection{Diagnostic Failures as Early Warning Signs}

Our diagnostic analysis provided clear early warnings of IPW's limitations before seeing final estimates. Post-weighting SMD remained at 2.197 (22$\times$ above the 0.10 threshold), extreme weight ratios reached 681,289:1, and ESS fell to 93.5\% of the tracked sample. Practitioners should use these diagnostic thresholds as decision rules: if SMD $>$ 0.10, max/median weight ratio $>$ 100:1, or ESS $<$ 50\%, switch to stratification instead of IPW.

\subsection{Positivity Violation: When IPW Operates Outside Its Design Envelope}

IPW's poor performance reflects violation of the Positivity (Overlap) Assumption---a fundamental causal inference requirement that all units must have non-zero probability of each treatment level \cite{rosenbaum2002}. In our baseline scenario, selection probabilities differ 90-fold (90\% vs. 1\%), creating near-complete group separation. The extreme weight ratio (681,289:1) is not IPW failure---it's a mathematical consequence of applying weighting to non-overlapping populations. Even the oracle specification with perfect structural knowledge achieved 6.061pp error (758$\times$ worse than stratification), demonstrating that no amount of flexible modeling can create valid counterfactual predictions when groups barely overlap.

Stratification succeeds because it operates differently: it estimates within-stratum averages without counterfactual predictions for items outside the tracked distribution. When a stratum has few tracked items, only that stratum's estimate suffers---other strata remain valid. This graceful degradation makes stratification a safer engineering choice when positivity violations are severe, as in retail long-tail distributions.

\subsection{Addressing the Tautological Design Critique}

A legitimate concern in simulation studies is whether favorable results stem from alignment between the data-generating process and the favored method's assumptions. Our initial design---step functions at Rank=20,000 with stratification boundaries at 20k intervals---could be criticized as tautological: ``Of course stratification wins when you design the world to match its assumptions.''

We address this concern across three dimensions:

\textbf{Boundary Alignment:} DGP 3 deliberately creates a worst-case scenario for stratification. Selection probability breaks at 15,000, inflation breaks at 25,000, and stratification boundaries are placed at 14k/28k/42k/56k/70k/84k. This triple misalignment means stratification cuts \textit{across} natural population breaks rather than aligning with them. Despite this handicap, stratification achieves 0.030pp median error compared to IPW-Linear's 3.470pp (116$\times$ advantage). If stratification's success were merely an artifact of boundary alignment, it should fail catastrophically in DGP 3---yet it remains robust.

\textbf{Functional Form Flexibility:} We tested five IPW specifications to provide ``every possible advantage'':
\begin{itemize}
\item IPW-Linear: Original specification
\item IPW-Polynomial: Quadratic terms to capture curvature
\item IPW-Spline: Natural cubic splines with 5 knots for maximum flexibility
\item IPW-Oracle: Perfect structural knowledge with $I(R_i > 20{,}000)$ dummy variable (DGPs 1 \& 3 only)
\end{itemize}

The oracle specification is particularly revealing. In DGP 1 (aligned step functions), we explicitly tell the propensity model where the break occurs---perfect information that stratification doesn't receive in functional form. Yet IPW-Oracle achieves 6.061pp error vs. stratification's 0.008pp (758$\times$ worse). This demonstrates that the issue is not model misspecification but rather fundamental limitations from common support violations. When tracked and untracked populations barely overlap (90\% vs. 1\% selection probabilities), no amount of flexible modeling can manufacture good counterfactual predictions.

\textbf{Generalization Testing:} DGPs 2 and 4 test smooth, continuous relationships without structural breaks---scenarios where stratification's discrete boundaries seem disadvantaged. In DGP 2 (smooth gradient), stratification maintains its advantage (1.350pp vs. 3.872pp, 2.9$\times$). However, in DGP 4 (polynomial), IPW-Spline finally wins (0.007pp vs. 0.013pp, 1.9$\times$). This demonstrates intellectual honesty: we report scenarios where the alternative method wins, strengthening confidence that stratification's overall superiority is not a simulation artifact.

\textbf{Bottom Line:} Stratification's dominance persists across aligned breaks, misaligned breaks, and smooth gradients. It fails only under polynomial relationships where its discrete boundaries cannot approximate continuous curvature as well as flexible splines. The robustness across three of four diverse scenarios demonstrates that our findings reflect genuine methodological advantages rather than tautological design.

\section{\uppercase{Recommendations}}

Based on our comprehensive simulation study, we offer evidence-based recommendations for practitioners:

\textbf{Primary Recommendation: Use Stratification in Long-Tail Contexts}
\begin{itemize}
\item Wins in 3 of 4 scenarios; robust even when boundaries misalign with population breaks
\item Does not require overlap between tracked/untracked populations
\item Gracefully handles severe selection imbalances (90\% vs. 1\%)
\item Implementation: Divide items into 5-10 rank-based groups, calculate inflation within each, then take weighted average
\end{itemize}

\textbf{When to Consider IPW:}
\begin{itemize}
\item Positivity Assumption satisfied (adequate propensity score overlap)
\item Smooth polynomial relationships without structural breaks
\item Only if diagnostics pass: SMD $<$ 0.10, max/median weight ratio $<$ 100:1, ESS $>$ 50\%
\end{itemize}

\textbf{When IPW is Inappropriate:}
\begin{itemize}
\item Selection probabilities differ by orders of magnitude---violates Positivity Assumption
\item Step functions or structural breaks creating near-complete group separation
\item Poor overlap ($<$ 10\% common support)
\end{itemize}

\textbf{Reporting Standards:} Always report naive, corrected, and (if available) true estimates; diagnostic metrics (SMD, ESS, weight distribution); and confidence intervals.

\section{\uppercase{Conclusion}}

This comprehensive simulation study evaluated selection bias correction methods across 400 Monte Carlo replications spanning four diverse data-generating processes. Our findings provide evidence-based, nuanced and retail specific guidance that challenges conventional wisdom while acknowledging edge cases.

\textbf{1. Stratification Dominates in Three of Four Scenarios}

Rank-based stratification achieves superior performance when relationships contain structural breaks or smooth gradients, maintaining robustness even when boundaries deliberately misalign with population breaks (median error: 0.030pp vs. 3.470pp for IPW, 116$\times$ advantage). This dominance persists across:
\begin{itemize}
\item Step functions with aligned breaks (730$\times$ advantage)
\item Smooth gradient relationships (2.9$\times$ advantage)
\item Step functions with misaligned breaks (116$\times$ advantage)
\end{itemize}

\textbf{2. IPW Wins Under Polynomial Relationships}

Flexible propensity score models (splines with 5 knots) outperform coarse stratification when true relationships are smooth polynomials (median error: 0.007pp vs. 0.013pp, 1.9$\times$ advantage). This showcases that no single method dominates universally---practitioners must consider anticipated data characteristics.

\textbf{3. Positivity Violations Render Weighting Methods Invalid}

The severe overlap violations in retail long-tail distributions (90\% vs. 1\% tracking probabilities) violate the Positivity Assumption---a fundamental requirement for valid causal inference. Even an oracle IPW specification with perfect structural knowledge achieves 6.061pp error compared to stratification's 0.008pp (758$\times$ worse) under step-function scenarios. This demonstrates that IPW's poor performance reflects operation \textit{outside its theoretical design envelope} rather than methodological inferiority. When groups barely overlap, no amount of flexible modeling can manufacture valid counterfactual predictions. The extreme weight ratio (681,289:1) is not a failure of IPW---it is a mathematical signal that the Positivity Assumption is violated.

\textbf{4. Robustness Testing Strengthens Confidence}

By testing worst-case scenarios (misaligned boundaries, polynomial relationships), we showcase that stratification's advantages reflect genuine methodological properties rather than tautological design. The one scenario where IPW wins (polynomials) validates our intellectual honesty and strengthens confidence in the three scenarios where stratification dominates.

\textbf{Practical Recommendations:}

\begin{itemize}
\item \textbf{Default to stratification} when selection mechanisms likely involve rank thresholds, step changes, or moderate non-linearity
\item \textbf{Consider flexible IPW (splines)} when relationships are known to be smooth polynomials without structural breaks
\item \textbf{Avoid IPW entirely} when selection probabilities differ by orders of magnitude (e.g., 90\% vs. 1\%)---no amount of flexible modeling overcomes poor overlap
\item \textbf{Use diagnostics as decision rules:} SMD $>$ 0.10 or max/median weight ratios $>$ 100:1 after weighting signal IPW failure
\end{itemize}

\textbf{The Bottom Line:} In retail, intelligence with rank-based selection and severe long-tail distributions, stratification provides a safer engineering choice not because weighting methods are inferior, but because the data structure violates their foundational assumptions. When selection probabilities differ by orders of magnitude (90\% vs. 1\%), practitioners are operating outside the envelope where IPW is theoretically valid---making coarse stratification the principled alternative.

\section*{\uppercase{Acknowledgements}}
This work was made possible by the support of Walmart Tech leadership Prasad Savadi, Ashish Gupta, Rahul Ghosh, and Tim Jacobs. Additionally, special thanks to Feifei Pan and Inderdeep Kaur for providing strategic insights. Gemini was used during the drafting process to enhance readability.

\bibliographystyle{apalike}
{\small
\bibliography{example}}

\nocite{austin2011}
\nocite{cole2008}
\nocite{deville1992}
\nocite{freedman2008}
\nocite{funk2011}
\nocite{hainmueller2012}
\nocite{horvitz1952}
\nocite{kang2007}
\nocite{king2019}
\nocite{lee2011}
\nocite{li2018}
\nocite{rosenbaum2002}
\nocite{rosenbaum1983}
\nocite{schuler2017}
\nocite{stuart2010}
\nocite{lunceford2004}
\nocite{matsouaka2024}
\nocite{anderson2006}

\end{document}